\documentclass[acmtog,authorversion,nonacm,screen]{acmart}

\usepackage{caption}
\usepackage{subcaption}
\usepackage{graphicx}
\usepackage{booktabs}
\usepackage{cleveref}
\usepackage{rotating}
\usepackage{float}
\begin{document}

\title{Towards Physically-Based Sky-Modeling}

\author{Ian J.~Maquignaz}
\orcid{0000-0002-6545-9029}
\affiliation{%
    \institution{Université Laval}
    \city{Québec}
    \state{Québec}
    \country{Canada}
}
\email{ian.maquignaz.1@ulaval.ca}

% \renewcommand{\shortauthors}{Maquignaz et al.}

%%%%%%%%% ABSTRACT %%%%%%%%%
\begin{abstract}

Accurate environment maps are a key component in rendering photorealistic outdoor scenes with coherent illumination. 
They enable captivating visual arts, immersive virtual reality and a wide range of engineering and scientific applications.
Recent works have extended sky-models to be more comprehensive and inclusive of cloud formations but existing approaches fall short in faithfully recreating key-characteristics in physically captured HDRI.
As we demonstrate, environment maps produced by sky-models do not relight scenes with the same tones, shadows, and illumination coherence as physically captured HDR imagery. 
Though the visual quality of DNN-generated LDR and HDR imagery has greatly progressed in recent years, we demonstrate this progress to be tangential to sky-modelling. 
Due to the Extended Dynamic Range (EDR) of 14EV required for outdoor environment maps inclusive of the sun, sky-modelling extends beyond the conventional paradigm of High Dynamic Range Imagery (HDRI).
In this work, we propose an all-weather sky-model, learning weathered-skies directly from physically captured HDR imagery.  
Per user-controlled positioning of the sun and cloud formations, our model (AllSky) allows for emulation of physically captured environment maps with improved retention of the Extended Dynamic Range (EDR) of the sky.
\end{abstract}

%% Keywords.

\begin{teaserfigure}
  \includegraphics[width=\textwidth]{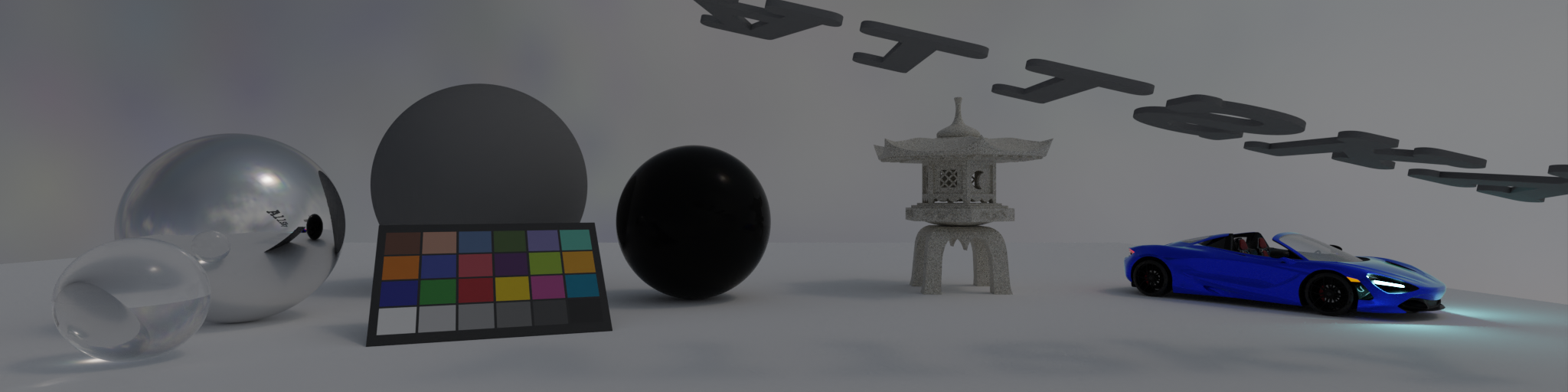}
  \caption{Blender rendered scene with an environment-map generated by AllSky per a hand-drawn user label}
  \label{fig:teaser}
\end{teaserfigure}

%%%%%%%%% TITLE %%%%%%%%%
\maketitle

%%%%%%%%% CONTENT %%%%%%%%%

\section{Introduction}

Outdoor illumination plays an important role in the human perception of physical spaces and the visual quality of media \& film \cite{CIE-History}.
Early works modelling skydomes combined data from varied sources into pre-computed and parametric sky-models for engineering and scientific applications, with the first sky-models \cite{MOON_1940,PEREZ_1993} modelling only luminance.
With the advent of the digital age, a new paradigm of digital-applications spurred renewed interest in sky-models.
In this regard, Nishita et al.~\cite{NISHITA_1993} proposed the first color sky-model enabling the generation of extraterrestrial views of the Earth for space-flight simulators, and Image-Based Lighting (IBL) techniques~\cite{IBL} were proposed to render synthetic objects into real and virtual scenes.

% Why use sky-models?
Though conventional low dynamic range (LDR) imagery can be suitable for some applications, High Dynamic Range (HDR) ~\cite{HDR_IBL_BOOK} imagery is integral to the photorealistic rendering of outdoor scenes.
% Though the definition of High Dynamic Range (HDR) is vague, 
Though HDR images broadly capture a larger range of illumination than what can be captured with conventional cameras, we note that the case of outdoor lighting is particular. 
Indeed, an estimated $22$ f-stops of exposure are necessary for HDR images to faithfully capture the highlights and shadows of an average real-world outdoor scene \cite{JENSEN_NightSky,HDR_IBL_BOOK,STUMPFEL_HDR_Sky_Capture}. 
For the purpose of clarity, we define Extended Dynamic Range (EDR) to describe images that capture the dynamic range of a reference scene with greater than conventional HDR dynamic range.
In this work we quantify dynamic range as exposure value (EV) given by $EV=log_2(|I|_{max} - |I|_{min})$, where $|I|$ is grayscale image intensity. 

% \todo{this paragraph is dangerous: you're using imagery that is inflexible etc. you're also not using parametric sky-models. You should mention this in the related work, or remove it altogether if already there.}
Physically-captured HDR images offer unsurpassed quality and photorealism in the rendering of skies, but physical capture is difficult, labour-intensive \cite{KIDER_captureFramework} and the resulting imagery is both inflexible and of fixed temporal- and geo-locality. 
As a means of mitigating these challenges, parametric sky-models have progressed greatly over the past four decades to offer a wide range of accurate, versatile, and portable models \cite{BRUNETON_2017_clearSky_eval} providing artists and creators a clear cost-advantage. 
Inclusive of both physically-based simulations and parametric models, sky-models are light-weight and computationally efficient, generating clear and overcast skies via a limited set of parameters. 
Though the limited set of user-parameters tied to physical properties are mostly-intuitive, these models do not parameterize atmospheric formations and are generally restricted to clear- and overcast-skies.
Popular models include the 4-parameter Hošek-Wilkie sky-model (HW \cite{HOSEK_13}), 
11-parameter Lalonde-Matthews outdoor illumination model (LM \cite{LM_2014}) and 4-parameter Prague Sky-Model (PSM \cite{PSM_21}).

In recent years, Deep Neural Networks (DNNs) have been proposed as all-encompassing sky-models capable of generating weathered skies learned directly from captured imagery.
These models vary in approach and per the conditionality (or lack thereof) available for users to control.
% Two of the earliest DNN sky-models allowed users to estimate an environment map for Image Based Lighting (IBL) from conventional LDR imagery \cite{YANNICK_2017, ZHANG_2019_panonet}, enabling accurate illumination, but visually limited to amorphous diffusion.
% Recent works often choose to augment an input HW \cite{DEEPCLOUDS_22}, LM \cite{LM-GAN_2023}, or PSM \cite{SKYGAN_22} parametric sky to an all-weather environment map with greatly improving on visual quality.
% Though some models offer generate clouds with unconditional stochasticity \cite{LM-GAN_2023, SKYGAN_22}, other such as DeepClouds \cite{DEEPCLOUDS_22} allows user control over the placement of cloud formations through binary-input masks.
% Most recently, Text2Light \cite{text2light} further improved on visual quality with the introduction of textually controlled generation of indoor and outdoor environment maps via diffusion.  

\begin{figure}[ht]
    \centering
    \includegraphics[width=\linewidth]{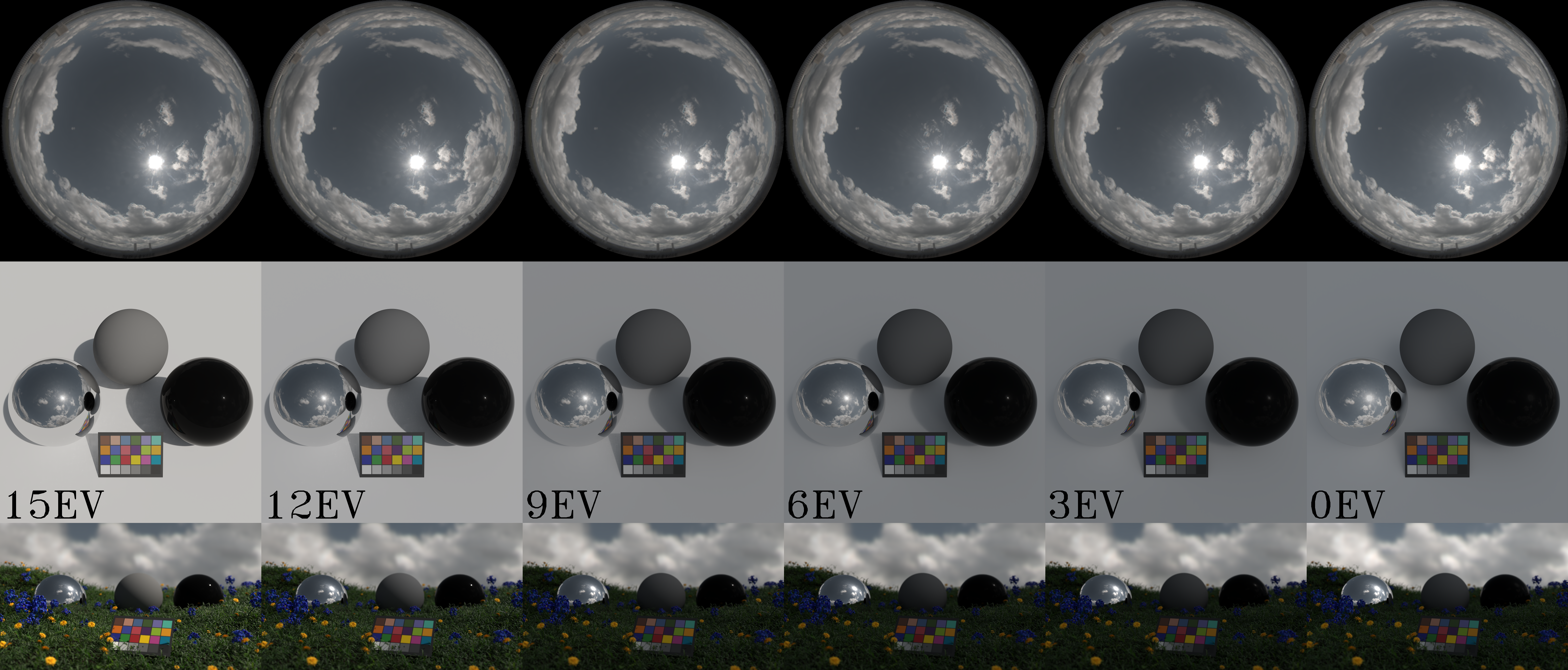}
    \caption{Impact of dynamic range on scene illumination. Each column illustrates the impact of an incremental clipping of dynamic range with exposure equalization to the 15EV ground truth.}
    \label{fig:3_demo_EV}
    \vspace{-0.5cm}
\end{figure}

Though there have been significant improvements in visual quality, the overall quality of an environment map and its illumination remains difficult to ascertain. 
As demonstrated by \cref{fig:3_demo_EV}, the incremental clipping of the dynamic range and equalizing of exposure with the 15EV ground truth results in images visually appearing unaltered. 
When rendered as a light source, the change in illumination energy becomes apparent resulting in softer tones, shadows, and light transmission. 

In this work, we target the generation of photorealistic environment maps matching the real-world luminance.
Our contribution is a conditional generative sky-model trained directly on physically captured HDR images without reliance on parametric sky-models.
Through our proposed pairing of intensity- and class-aware Selective and Cascade losses with ANN LDR2HDR booster, we demonstrate generated environment maps with greater retention of dynamic range and illumination.
User-friendly input modalities allow user-control over the placement of the sun and cloud formations, enabling the rendering of photorealistic outdoor scenes with coherent illumination: achieving higher accuracy in rendered tones, shadows, and light transmission.

% \todo{Describe the main contributions: losses, latent2hdr? I commented out this last sentence as it's unclear... rephrase?}

% Our method neither augments nor generalizes to a parametric sky-model, and our model directly outputs HDR environment maps without the need for further decompression.
\section{Background}

% Parametric and physically-based models
To limit computational requirements, it is often preferable to approximate physically-based simulations of solar and atmospheric illumination via mathematical models \cite{KIDER_captureFramework}.
One formulation is through the development of numerical models such as
~\cite{NISHITA_1993,NISHITA_1996,ONEAL_2005,HABER_2005,BRUNETON_2008,ELEK_2010}, which derive simplified mathematical representations for complex atmospheric systems.
Though reducing computation expense, numerical models generally remain complex, memory intensive, and require a pre-computation step.
An alternative formulation introduced by Perez et al.~\cite{PEREZ_1993} is the fitting of an analytical model to a body of sky data.
These simpler models can be fitted to computationally expensive, accurate, and diverse data acquired from complex models (e.g., Preetham et al.~\cite{PREETHAM_1999} is fitted to \cite{NISHITA_1996}), \cite{HOSEK_13,HOSEK_13Sun,PSM_21} use path tracers such as A.R.T \cite{ART}, and Bruneton \cite{BRUNETON_2017_clearSky_eval} evaluated against libRadtran \cite{libRadtran} physical simulations and Kider \cite{KIDER_captureFramework} physically captured dataset.
Such models trade-off accuracy to produce versatile, lightweight, flexible, and fast parametric models which support a wide range of applications through a finite set of user-parameters \cite{BRUNETON_2017_clearSky_eval}.

% Deep Learning
With the advent of deep learning, the fitting of analytical models to features and modalities can be automated, with parameterization to a finite set of latent parameters, or to intuitive user-parameters.
Notably, this concept has been proposed for lighting estimation~\cite{YANNICK_2017, YANNICK_2019_SKYNET, ZHANG_2019_panonet, relighting_hdsky, Yu2021DualAA}, where LDR images can be used to guide the generation of HDR environment maps for relighting virtual objects and scenes.
Though these learned representations tend to offer little photorealism and generate overly smooth skies, they accurately capture illumination and enable a wide range of precision-oriented applications.

Recent deep learning approaches have been proposed to augment parametric clear-skies with photorealistic clouds.
CloudNet \cite{DEEPCLOUDS_22} proposed a method for augmenting Ho{\v{s}}ek-Wilkie~\cite{HOSEK_13} parametric skies per user-controlled cloud placement.
SkyGAN~\cite{SKYGAN_2022} proposed fitting the Ho{\v{s}}ek-Wilkie~\cite{PSM_21} model to real-world photographs, enabling the re-generation of Ho{\v{s}}ek-Wilkie skies with cloud formations, though with limited photorealism.
This formulation was also demonstrated by LM-GAN \cite{LM_GAN_2023}, fitting physical captures to the 11-parameter Lalonde-Matthews (LM) outdoor lighting model \cite{LM_2014} and producing photorealistic, though lower-resolution, weathered skies.
Most recently, Text2Light \cite{text2light} introduced the textually controlled generation of indoor and outdoor environment maps via diffusion. 
Though this model offers unprecedented LDR visual quality, the textual conditioning leaves a lot to be desired. 

% Shortfalls to current methods
Physically-accurate sky-models model illumination of clear, hazy, and/or overcast daylight skies with performance evaluated against physically captured data, but modelling and control over of non-uniform atmospheric formations such as clouds remains a challenge.
Many works propose augmenting clear-sky-models with cloud formations, but this does not retain a tractable relationship with the finite parameters of the underlying physically-accurate clear-sky-model.
As a result, augmented environment maps generally exhibit inconsistent and depressed illumination. 
To mitigate the problem, many works propose post-generation tricks to improve illumination (particularly of the sun), including aggregating a parametric sun model and further mathematical decompression \cite{SKYGAN_2022, DEEPCLOUDS_22, text2light}.
These methods offer varied photorealism and illumination accuracy. 
A common alternative is to include complex cloud formations through volumetric cloud rendering~\cite{DeepScatering} and cloud simulations~\cite{BRUNETON_2008_clouds} post environment-map generation.
While this multi-step approach can be versatile and photorealistic, it can be labour intensive, computationally expensive and difficult to configure \cite{SIGGRAPH_course_2020}.

As a result of current complications, the modelling of weathered outdoor skies remains a challenge without a clear solution \cite{Apple_Paper}.
Physical capture remains unsurpassed and often preferable for both photorealism and weathering \cite{HOSEK_13Extrasolar,FORZA}.

\section{Methodology}

\subsection{Dynamic Range}
\begin{figure}[ht]
  \centering
    \vspace{-0.4cm}
  
\begin{subfigure}{.95\linewidth}
  \centering
  \includegraphics[width=\linewidth]{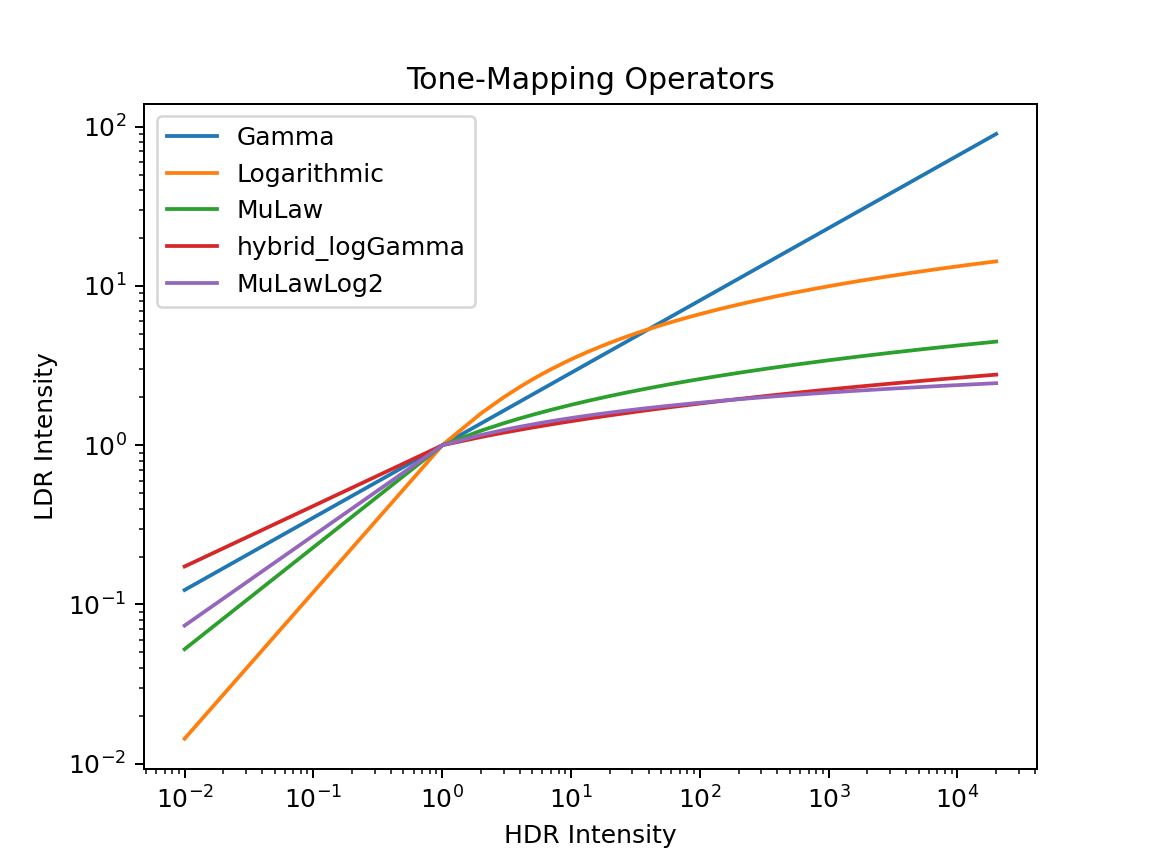}
  \vspace{-0.5cm}
  \caption{Tone-Mapping HDR intensity}
  \label{fig:3_plt_tm}
\end{subfigure}
\hfill
\begin{subfigure}{.95\linewidth}
  \centering
  \includegraphics[width=\linewidth]{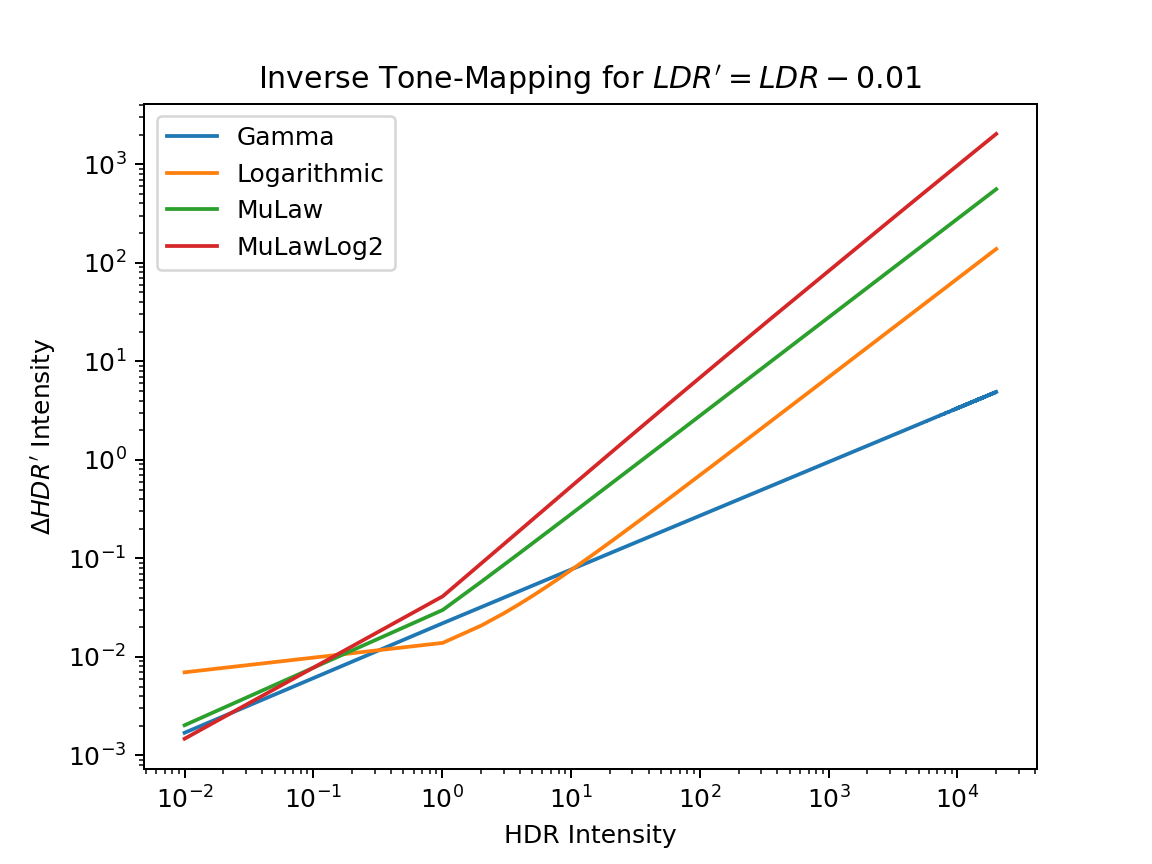}
  \vspace{-0.5cm}
  \caption{Impact of LDR error $\delta-=0.01$ on HDR space}
  \label{fig:3_plt_tm_error}
\end{subfigure}
\caption{Tone-Mapping Operators}
\vspace{-0.5cm}
\end{figure}

Tone-mapping operators are commonplace to compress Dynamic Range (DR) to a visible (or latent) color-space more favourable for DNN training. 
We investigate a range of operators ($\tau$), including
logarithmic ($Log_n$) $I' = \log_n(I+1)$,  
Power-Law ($\gamma$) $I' = I^{\frac{1}{\gamma}}$, 
$\mu$-law $I' = \frac{\log_e(1+\mu I)}{\log_e(1.0+\mu)}$,
and combinations thereof as shown in \cref{fig:3_plt_tm}. 
Each operator is a bijection, allowing for recovery of the original image via $I = \tau^{-1}(I')$ but, as shown in \cref{fig:3_plt_tm_error}, these operators introduce non-linearity between error $\delta$ in LDR compressed space and error $\Delta$ in uncompressed HDR space.

Given the emerging adoption of mixed-tonemappers, tonemappers comprised of combinations of tonemappers, such as Hybrid Log-Gamma, we theorize that an optimal tonemapper for DNN training restricts values to a favourable dynamic range such as $-1 \lessapprox x \lessapprox 1$. 
In this regard, we propose a hybrid $\mu$-lawLog$_2$ tonemapper as defined in \cref{eq:muLawLog2} and illustrate in \cref{fig:3_plt_tm}.
We propose this tonemapper to better preserve texture-rich clouds and skydome components (primarily $EV \lessapprox 1$) and aggressively compress texture-poor high intensity regions which are primarily saturated in LDR imagery.

\begin{equation}
\mu\text{-lawLog}_2\left(I\right) = \log_2\left[ \frac{\log_e(1+\mu I)}{\log_e(1.0+\mu)} +1\right]
\label{eq:muLawLog2}
\end{equation}

\subsubsection{Losses}
In experimentation, we observed that an $L_1$-loss on a dataset of clear-sky imagery guides a model towards a median clear-sky image with few visual artifacts, but IBL renderings exhibit grossly different scene illumination.
We associate this discrepancy with lost dynamic range, particularly in solar regions. 
To mitigate, we propose exposure- and segmentation-aware losses. 

\begin{figure*}[ht]
    \centering
    \subfloat[3D Surface of a Skydome's intensity (EV)] {
        \includegraphics[width=0.35\linewidth,keepaspectratio]{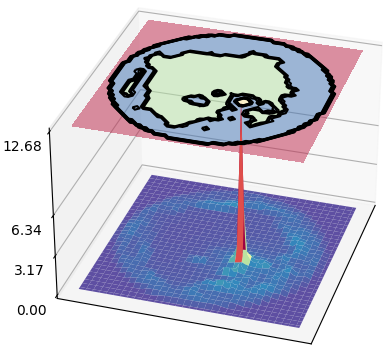}
        \label{fig:3_plt_3D_surface}
    } 
    % \hfill
    \subfloat[Loss sensitivity to truncation of environment-map dynamic range (EV) in HDR-space (left) and LDR-space (right). LDR-space illumination ratio presents retained illumination from ground truth environment map after dynamic range truncation. $\alpha(e)=1$] {
        \includegraphics[width=0.55\linewidth,keepaspectratio]{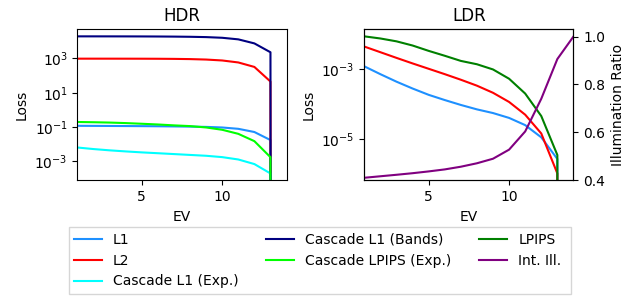}
        \label{fig:3_plt_loss_sensitivity}
    }
    \vspace{0.2cm}
    \caption{Database sample, June 7th, 2016 at 1:54PM}
\end{figure*}

As illustrated in \cref{fig:3_plt_loss_sensitivity}, the conventional global $L_1$-loss is insensitive to the dynamic range of the sun for HDRI clamped to increasing EV. 
Though an $L_2$-loss offers greater sensitivity, its application in the HDR-space is likely to result in a loss-imbalance and/or a gradient explosion.

An impeding factor in conditioning is the class-imbalance exhibited by outdoor environments maps. 
As shown by the sample HDRI in \cref{fig:3_plt_3D_surface}, sun pixels represent a small subset of pixels (\textcolor{red}{red}; 3.6\% of pixels and $EV=13.4$ in this sample) with intensity that belittles the remainder of the skydome (\textcolor{blue}{blue}). 
In computing integrated illumination (\cref{eq:IntegratedIllumination}, plotted in \cref{fig:3_plt_loss_sensitivity} as Int. II), the segmented solar region represents 61\% of the environment maps' illumination and the skydome the remaining 39\%. 

To mitigate this loss in illumination we propose exposure-aware `Cascade' losses, which segment HDRI by exposure brackets. 
Where dimensionality is impertinent, \cref{eq:cascade_bands} performs a pixel-wise selection for each exposure bracket via the mask computed in \cref{eq:cascade_bands:mask} from $I_r$.
We apply this concept to reconstruction losses (e.g.\ $L_1$) via substitution of $\mathcal{L}_f$, effectively limiting the sample count to a selected subset of pixels. 
Note, we floor exposure such that $2^{-1}=0$ and $\alpha(x)$ is an optional adaptive-scaling factor.

\begin{equation}
    M = 2^{i-1} \leq I_{real} \leq 2^{i}
    \label{eq:cascade_bands:mask}
\end{equation}
\begin{equation}
\mathcal{L}_{bands}(I_{r},I_{f}) = \sum_{i=0}^{i=EV_{max}} \alpha(x)\mathcal{L}_f(I_{r}\subset M, I_{f}\subset M) 
\label{eq:cascade_bands}
\end{equation}

Where dimensionality is pertinent, \cref{eq:cascade_exposure} re-exposes the input HDRI for each exposure bracket. 
This enables the substitution of $\mathcal{L}_f$ with visual losses (e.g.\ LPIPS) with the optional application of class-selection via mask (M), $\gamma$-tonemapping and clamping at each exposure.

\begin{equation}
\mathcal{L}_{cascade}(I_{r},I_{f}) = \sum_{i=0}^{i=EV_{max}} \alpha(x) \mathcal{L}_f({2^{-i}}I_{r}\cdot M, {2^{-i}}I_{f}\cdot M) 
\label{eq:cascade_exposure}
\end{equation}

\begin{equation}
\mathcal{L}_{f}(I_{r},I_{f}, M) = \mathcal{L}_f(I_{r}\subset M,I_{f} \subset M) 
\label{eq:selective}
\end{equation}

We further propose the implementation of class-segmented losses, allowing conventional metrics (e.g.\ L$_1$) and our proposed Cascade metrics to be selectively applied via a segmentation mask $M$ per \cref{eq:selective}.
This segmentation allows losses to distinguish between classes which may have varying dynamic ranges, or classes to be tuned exclusively by favourable losses. 
Our proposed segmentation specifically targets lost sensitivity in specific domains, such as the border, skydome or sun, which would otherwise be saturated by pixels from other domains. 

\subsection{Model}

\begin{figure}[ht]
    \centering
    \vspace{-0.3cm}
    \subfloat {
      \includegraphics[width=\linewidth]{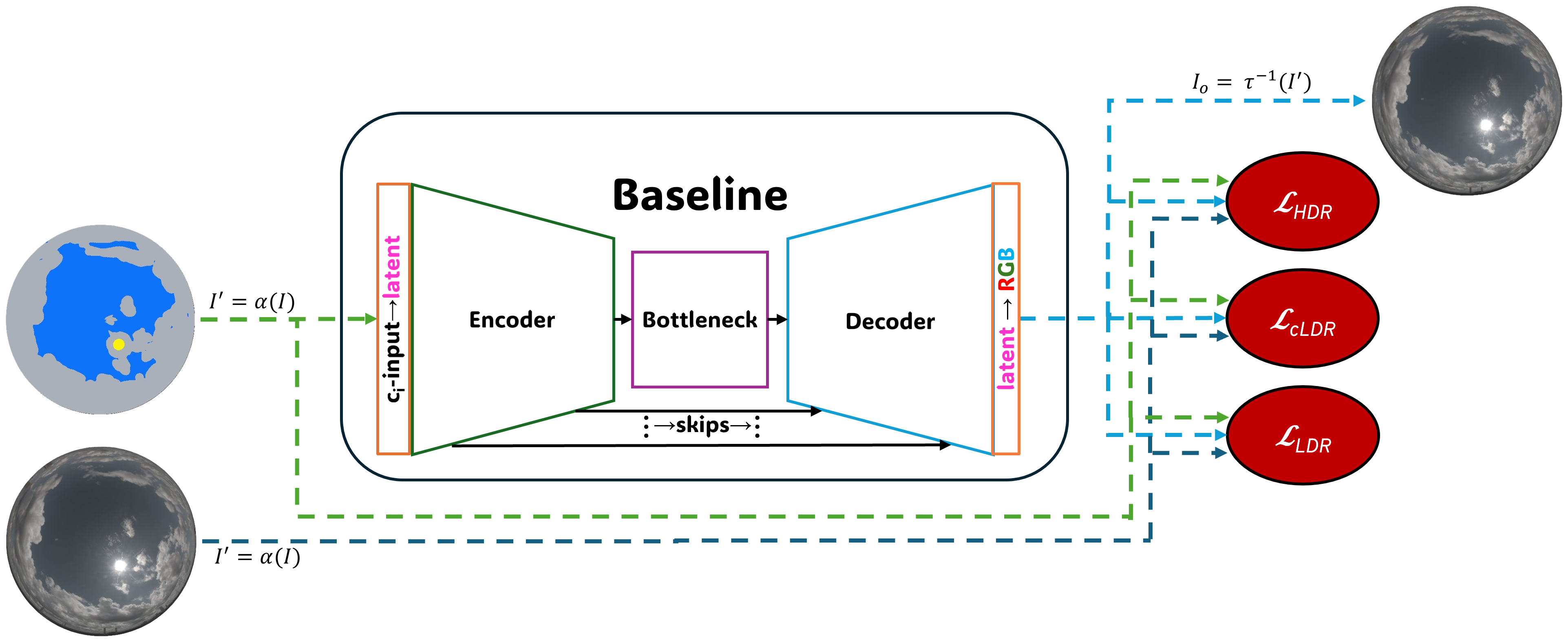}
    } \hfill
    \vspace{-0.5cm}
    \subfloat {
      \includegraphics[width=\linewidth]{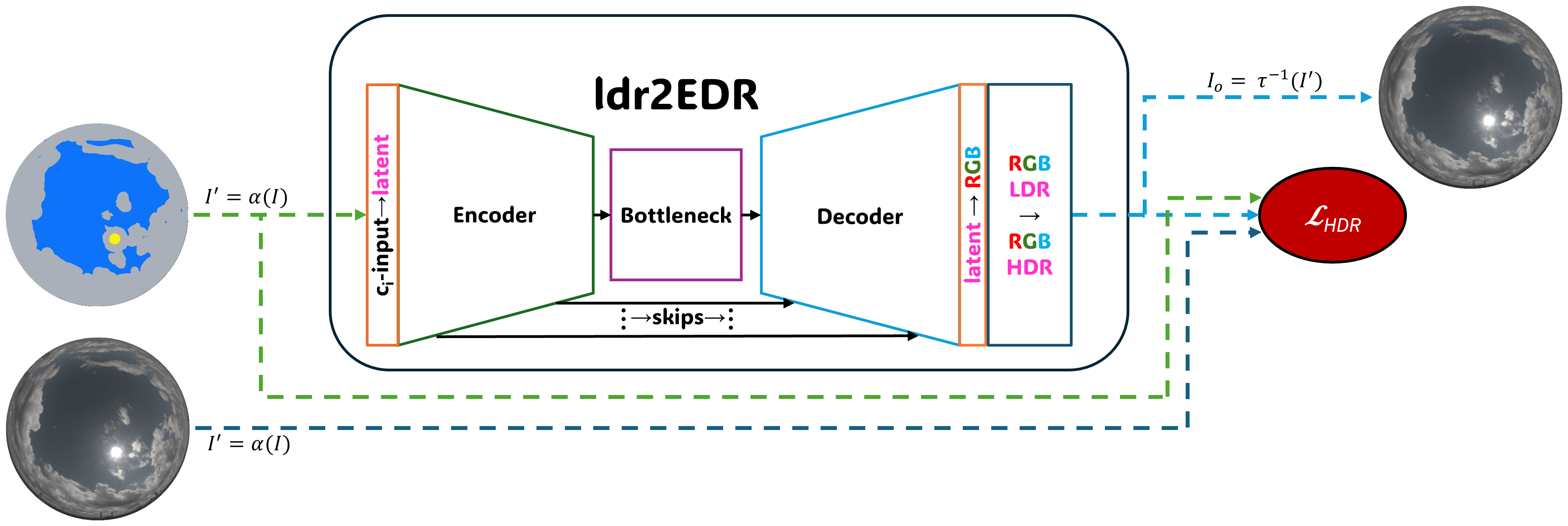}
    } \hfill
    \vspace{-0.5cm}
    \subfloat {
      \includegraphics[width=\linewidth]{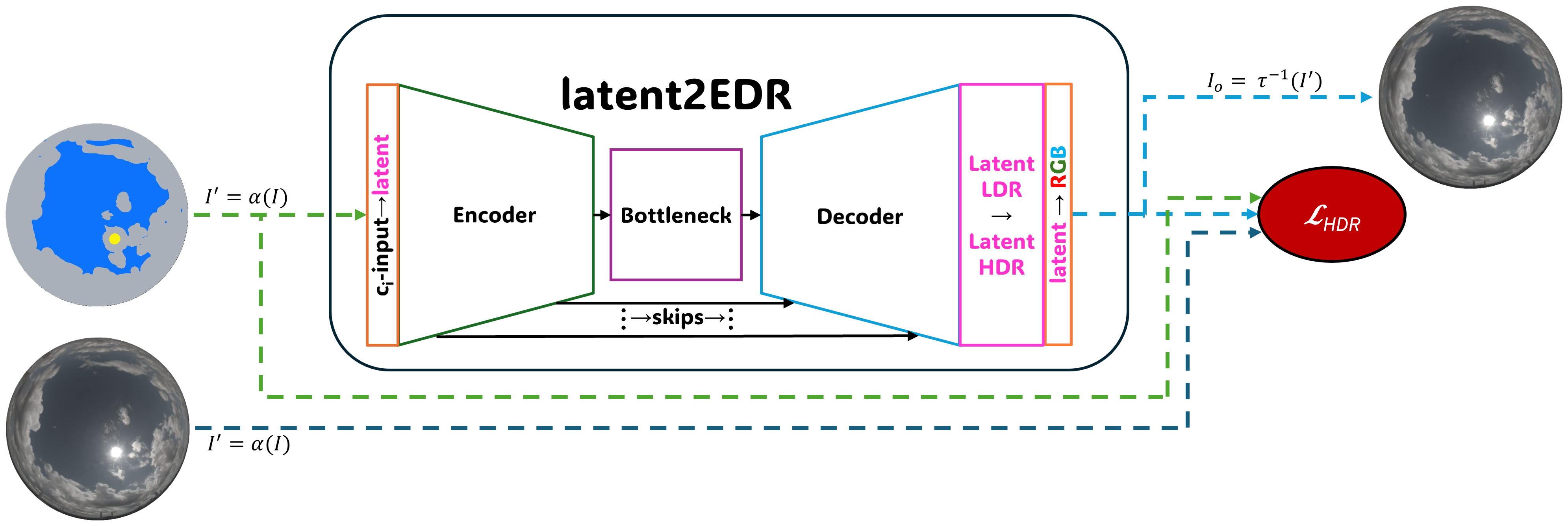}
    }
    \caption{AllSky}
    \label{fig:3_dia_AllSky}
    \vspace{-0.65cm}
\end{figure}

We propose organizing sky-models metrics per the \textit{Baseline} in \cref{fig:3_dia_AllSky}, evaluating output as $\mathcal{L}_{LDR}$ compressed HDRI, $\mathcal{L}_{cLDR}$ compressed and clipped HDRI (visual losses), and $\mathcal{L}_{HDR}$ EDR HDRI. 
In training, we propose the selective application of losses to each group to allow dynamic-range-aware guidance. 

We recognize the emphasis of tonemapping literature is the compression of HDRI to a human-visible/displayable format which is presumed to carry-forward to DNN training. 
We hypothesize that these domains, human-visible/displayable and DNN training, are distinct and the aforementioned tonemappers are a priori. 
If the human-visible requirement is removed, the objective becomes grossly undefined. 
Given a tonemapper is a single or combination of non-linear functions, we hypothesize that a tonemapper can be approximated as a piecewise linear spline function learned by an ANN.

Given the current paradigm, we theorize that DNNs trained in compressed-space are ignorant of the non-linear relationship between LDR and EDR space, and thus learn a depressed-LDR colorspace due to a presumption of constant error importance. 
To correct this, we propose an ANN-tonemapper which takes RGB values in depressed-LDR colorspace and learns an inverse-tonemapping function to physically-accurate RGB-EDR. 
We illustrate this model in as \textit{ldr2EDR} in \cref{fig:3_dia_AllSky}, taking RGB-LDR model output and mapping it to EDR.
In this configuration, the backbone model can be trained using one of the aforementioned tonemappers prior to training with the \textit{ldr2EDR} ANN.

We further theorize that the collapse of DNN latent colorspace to RGB represents a loss of information which may be of benefit to an ANN-tonemapper.
As such, we propose \textit{latent2EDR} in \cref{fig:3_dia_AllSky}, which mitigates collapsing the U-Net's latent color-space to RGB prior to ANN decompression to RGB-EDR.
In this configuration, the backbone model with RGB output layer can be trained using one of the aforementioned tonemappers, then substituted with a latent n-channel output layer and the \textit{latent2EDR} ANN.

\section{Experiments}

\subsection{Evaluation methodology}

For the purpose of visualization, all images in this work are Gamma ($\gamma=2.2$) tonemapped.

\subsubsection{Dataset}
\begin{figure}[ht]
    \centering
    \begin{subfigure}{.45\linewidth}
        \centering
        \includegraphics[width=\linewidth]{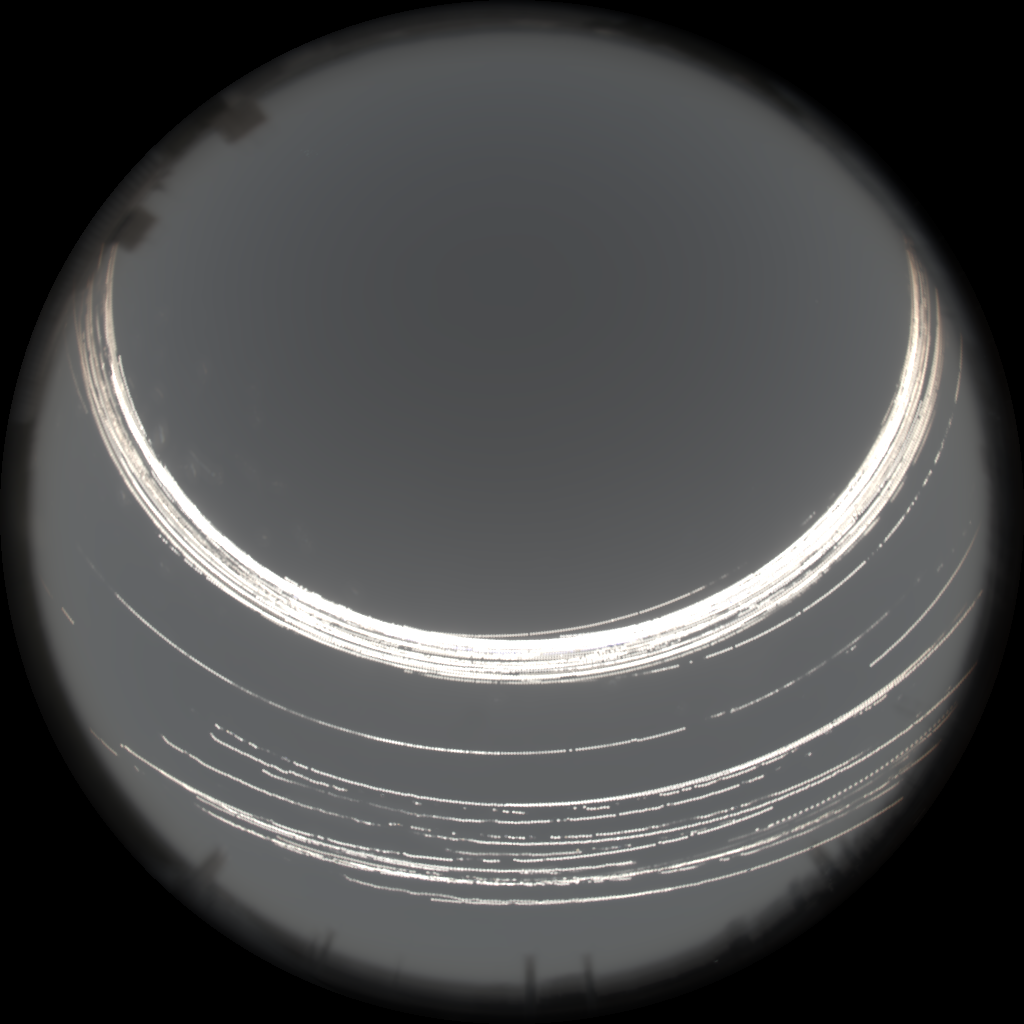}
        \caption{Mean Skydome}
        \label{fig:3_img_mean_skydome}
    \end{subfigure}
    \begin{subfigure}{.45\linewidth}
        \centering
        \includegraphics[width=\linewidth]{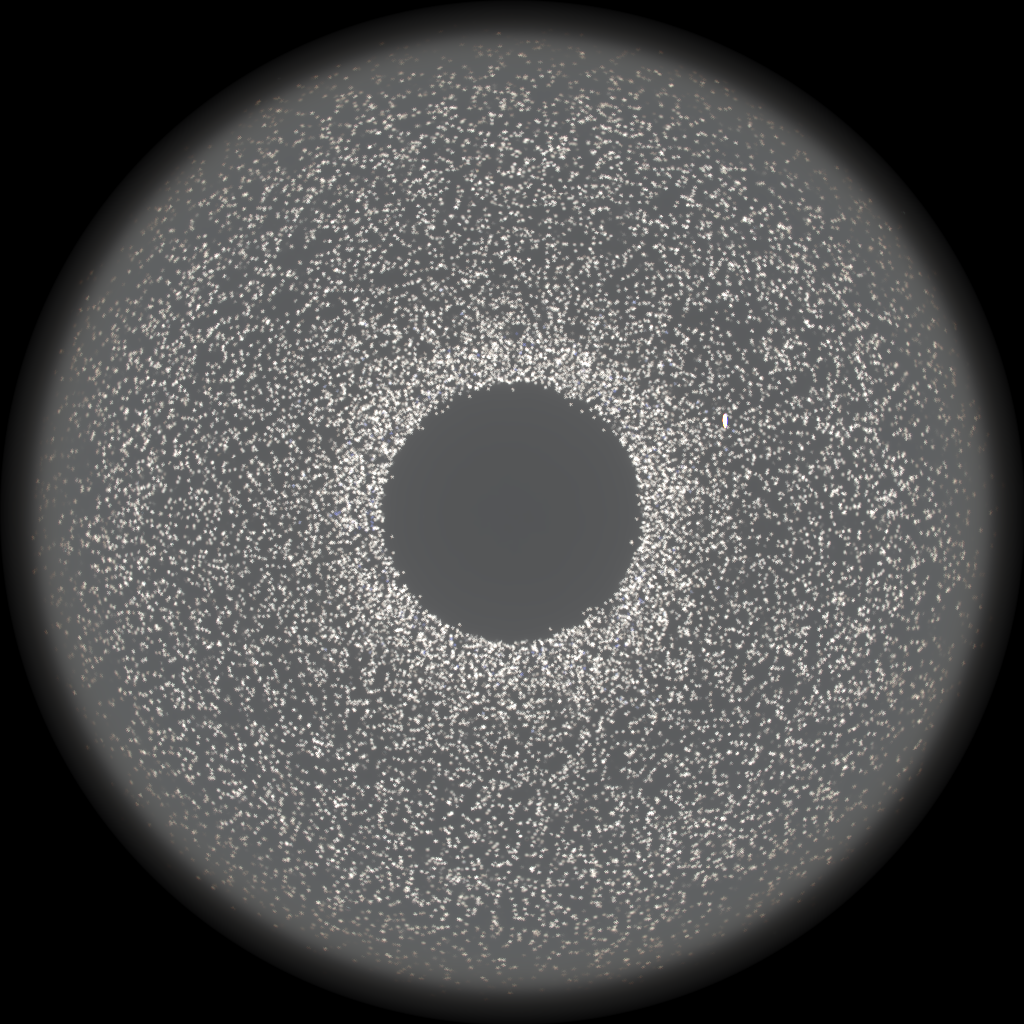}
        \caption{Mean Augmentation}
        \label{fig:3_img_mean_augmented_skydome}
    \end{subfigure}
    \caption{HDRDB Skydomes}
    \vspace{-0.3cm}
\end{figure}

The Laval HDR Sky database (HDRDB, \cite{LavalHDRdb}) consists of 34K+ HDR images captured in Quebec City, Canada at varied intervals between 2014 and 2016 using the capture method proposed by Stumpfel et al.~\cite{STUMPFEL_HDR_Sky_Capture}.

% \begin{figure}[h]
%     \centering
%     \includegraphics[width=\linewidth]{figures/fig_skyangular.pdf}
%     \caption{Sky-Angular Environment Map}
%     \label{fig:3_dia_skyangular}
% \end{figure}

From the database—the mean of which is shown in \cref{fig:3_img_mean_augmented_skydome}—we augment the dataset with random rotations around the zenith to increase solar placement coverage (\cref{fig:3_img_mean_augmented_skydome}) and enable generation of skies outside of HDRDB.
We split the dataset into training (25,243), validation (3,697), and testing (3,519) subsets by arbitrarily splitting by date of capture such that each subset has a random assortment of images from each year and season of capture.

\subsubsection{Image-Processing}
We develop a pre-processing pipeline to downsample and augment, optimizing for retention of physically captured characteristics. 
This minimized unwanted alteration of Dynamic Range (EV), color-space, and Integrated Illumination ($\oiint_I$). 
We achieve this by augmenting with linear spline interpolation prior to downsampling and downsampling by factors of $f(x)=1/2^x$ to enable inter-area interpolation (retaining over 99\% of illumination energy).
For evaluation, we quantify pixel intensity as linear luminance per BT.709 \cite{BT709} and further quantify global-illumination as Integrated Illumination:
\begin{equation}
    \oiint_I(I) = \sum{\Omega|I|}
    \label{eq:IntegratedIllumination}
\end{equation} 
Where $\Omega$ is the environment-map's solid angles (pixel-wise angular field-of-view).

\subsubsection{Segmentation}
Solar positioning was identified via ephemeris calculations \cite{pySolar}.
From an extraterrestrial perspective, the solar disc is 0.5$^{\circ}$ in diameter which we extend to a masked solar-region 5$^{\circ}$ to include the solar corona and to reflect atmospheric attenuation of solar radiance.

Segmentation of cloud formations was achieved by thresholding the ratio  $Y = \frac{B-R}{B+R}$ proposed by Dev et al.~\cite{clouds_segmentation}. 
We observe that $\mu$LawLog$_2$ tonemapping the HDRI visually improves segmentation, finding the segmentation is robust but variable under different lighting and seasonality. 
Clouds masks are further processed morphologically to emulate creation using a parametric circular brush of 15 pixels. 
These operations are performed to improve boundaries, mitigate overfitting to the segmentation, and lastly, emulate `hand-drawn' masks to improve model usability.

The final label is produced by combining solar, cloud formation, border and skydome (clear-sky) regions into a uniform 1-channel label. 

\subsection{Baseline Models}

\subsubsection{Baseline}
As a baseline and backbone, we implement UNetFixUp \cite{UNetFixup} as a proven model for outdoor image editing which can be configured for arbitrary $c_{i}$-channels discrete ($x\in\mathbb{Z}$) or continuous ($x\in\mathbb{R}$) input and $c_{o}$-channels output. 
We illustrate this configurability in \cref{fig:3_dia_AllSky} by emphasizing the input and output layers to the U-Net architecture.

\subsubsection{AllSky: ldr2EDR \& latent2EDR}

The ANN models ldr2EDR and latent2EDR are implemented as variable number of ReLU activated convolutional layers with kernel size $k=1$. 
Functionally an MLP, this approach mitigates a requirement to alter the dimensionality of the data and improves performance. 

Unless otherwise specified, we define the following losses:
\begin{enumerate}
    \item LDR: Per-class selective $L_1$ loss for border, skydome and sun with $\alpha$-sun$=0.1$
    \item cLDR: Global LPIPS loss
    \item HDR: Per-class Selective $L_1$ loss for border and skydome with $\alpha$-skydome$=10$, global Cascade LPIPS with 4 exposures, and Selective Cascade with 15 Bands on the solar region with $\alpha$-sun$=0.1$
\end{enumerate}

In addition to the RGB or latent output of the backbone, we propose providing a priori of $X,Y,Z,\Omega, Label$.

\subsubsection{DeepClouds} 
DeepClouds was reproduced per the authors' source-code with model and parameters unchanged \cite{DEEPCLOUDS_22}.
We adapt the data-processing pipeline to our segmentation of HDRDB, reproducing the input histogram-equalized Ho\v{s}ek-Wilkie \cite{HOSEK_13} clear-skies, cloud distance field maps and tonemapping (\cref{eq:DeepClouds_compression}).
Originally trained on a small dataset of 629 HDRI for 7,500 epochs, we implemented a pytorch-lightning framework to facilitate training of DeepClouds on the $40$x larger HDRDB dataset and completed a prolonged 3,500 epoch training.

\begin{equation}
I' = 1 / \left(1 + I + 0.01\right)
% I = (1 / I') - 1.01
\label{eq:DeepClouds_compression}
\end{equation}

We evaluate DeepClouds both with and without the sun-pass-through mechanism by repeating the model's evaluation while toggling the summation of the Ho\v{s}ek-Wilkie clear-sky solar disk into the generated environment map.
To evaluate the input modalities, we evaluate a modified baseline of DeepClouds with our 1-channel label as input.

\subsubsection{Text2Light}
Evaluation of Text2Light's \textit{outdoor} functionality was completed via unaltered implementation the author's published model, checkpoints and textual CLIP prompts with no upscaling ($\beta=1$) \cite{text2light}.
Generated LDR and HDR imagery were matched to ground truth images from the Laval Outdoor Dataset (LOD) \cite{YANNICK_2019_SKYNET}. 
Given visual similarity and negligible misalignment, this was accomplished by FLANN matched ORB keypoints and a scale invariant loss (\cref{eq:scale_invariant_loss}\cite{eigen2014depth}).

\begin{equation}
    \mathcal{L}(I_{r},I_{f}) = \frac{1}{n}\sum (\ln{I_{r}} - \ln{I_{f}})^2 - \frac{1}{n^2} \left( \sum (\ln{I_{r}} - \ln{I_{f}}) \right)^2
    \label{eq:scale_invariant_loss}
\end{equation}

% \begin{equation}
%     \alpha \frac{1}{MN}\sum_{i=0}^{M}\sum_{j=0}^{N} \Omega |I_{ij}| = \frac{1}{MN} \sum_{i=0}^{N}\sum_{j=0}^{N} \Omega|I_{ij}|
% \end{equation}

The 80 image pairs were normalized for exposure and compared for visual metrics as skyangular environment maps. 
Text2Light's inverse ToneMapping Operator (iTMO) MLP is trained on tonemapped LOD HDRI using a linear-scale invariant loss (\cref{eq:scale_invariant_loss}).
Thus for a given HDRI from LOD, a constant $\alpha$ should exist such that $|I_{real}| = \alpha |I_{fake}|$. 

\subsubsection{Metrics}

The objective of this work is to illustrate the importance of dynamic range and illumination, as well as the implications of their misrepresentation in downstream applications such as IBL.   
We target EDR (EV) and illumination ($\oiint_I$) to evaluate environment maps, quantitatively presented in relation to the ground truth as ratios of the average of generated ($I_f$) and ground truth ($I_r$) HDRI for dynamic range (EV ${I_f}/{I_r}$) and illumination ($\oiint_I$ ${I_f}/{I_r}$). 
To demonstrate the importance and sensitivity of these metrics, we include $L_1$ and $L2$ for LDR- and HDR-space. 

Given most textures are lower exposure brackets, we assume visual quality metrics are insensitive to tonemapping and include LPIPS \cite{LPIPS}, CLIP-IQA \cite{CLIP_IQA}, and FID \cite{FID} to demonstrate retention of visual quality. 
For the purpose of uniform evaluation, all generated imagery is Gamma ($\gamma=2.2$) tone-mapped to clipped LDR-space for visual metrics.

\subsection{Comparison}

\begin{table*}[htb]
    \caption{DeepClouds, U-Net with clear-sky priori}
    \label{tab:clearSky}
    \begin{tabular}{@{}rcccccccccc@{}}
    \toprule
    & LDR $L_1$ $\downarrow$ & LDR $L_2$ $\downarrow$ & HDR $L_1$ $\downarrow$ & HDR $L_2$ $\downarrow$ & LPIPS $\downarrow$ & CLIP-IQA $\uparrow$ & EV ${I_f}/{I_r}$ & $\oiint_I$ ${I_f}/{I_r}$ \\
    \toprule
     DeepClouds               & 0.013 & 0.0008 & 0.07 & 412 & 0.21 & 0.54 & 0.36 & 0.59 \\
     DeepClouds w/ Sun        & 0.019 & 0.006  & inf  & inf & 0.21 & 0.53 & 0.36  & $3e^{31}$ \\
     DeepClouds w/o Clear-Sky & 0.016 & 0.0012 & 0.07 & 417 & 0.21 & 0.54 & 0.36 & 0.62 \\
    \bottomrule
    \end{tabular}

    \vspace{5pt}
    \caption{Tone-mapper Ablation. Ground Truth CLIP-IQA is 0.36}
    \label{tab:fixUpUnet_tonemappers}
    \begin{tabular}{@{}rcccccccccc@{}}
    \toprule
    Tone-mapper & LDR $L_1$ $\downarrow$ & LDR $L_2$ $\downarrow$ & HDR $L_1$ $\downarrow$ & HDR $L_2$ $\downarrow$ & LPIPS $\downarrow$ & CLIP-IQA $\uparrow$ & EV ${I_f}/{I_r}$ & $\oiint_I$ ${I_f}/{I_r}$ \\
    \toprule
    %1 OK 112 --model_name=FixUpUnet --mode_tonemap=GAMMA --batch_size=2 --loss_no_LDR=False --loss_no_cLDR=False --loss_no_HDR=True --lr_G=0.00001
    %2 OK 86  --model_name=FixUpUnet --mode_tonemap=LOG2  --batch_size=2 --loss_no_LDR=False --loss_no_cLDR=False --loss_no_HDR=True --lr_G=0.00001
    %3 OK 114  --model_name=FixUpUnet --mode_tonemap=MIXED --batch_size=2 --loss_no_LDR=False --loss_no_cLDR=False --loss_no_HDR=True --lr_G=0.00001
    $\gamma$              & 0.057 & 0.031  & -     & -      & 0.25 & 0.36 & 0.19  & 0.48 \\
    $Log_2$               & \textbf{0.047} & \textbf{0.009}  & 0.087 & 428.48 & 0.28 & 0.36 & \textbf{0.27} & \textbf{0.58} \\
    $\mu\text{-lawLog}_2$ & 0.058 & 0.010  & 0.078 & \textbf{422.76} & \textbf{0.23} & 0.36 & \textbf{0.27} & \textbf{0.58} \\
    
    %4 OK 112 --model_name=FixUpUnet --mode_tonemap=MIXED --batch_size=2 --loss_no_LDR=False --loss_no_cLDR=False --loss_no_HDR=True --lr_G=0.00001 --loss_LDR_global_L1 --loss_HDR_global_L1
    %5 OK 93  --model_name=FixUpUnet --mode_tonemap=MIXED --batch_size=2 --loss_no_LDR=True --loss_no_cLDR=False --loss_no_HDR=False --lr_G=0.00001 --loss_LDR_global_L1 --loss_HDR_global_L1
    $\mu\text{-lawLog}_2$ Global LDR $L_1$ & 0.055 & \textbf{0.009} & \textbf{0.077} & 423.15 & \textbf{0.23} & 0.36 & 0.14 & 0.52 \\
    $\mu\text{-lawLog}_2$ Global HDR $L_1$ & 0.056 & \textbf{0.009} & \textbf{0.077} & 422.96 & \textbf{0.23} & 0.36 & 0.19 & 0.54 \\
     
    %6 FAILED --model_name=FixUpUnet --mode_tonemap=GAMMA --batch_size=2 --loss_no_LDR=True --loss_no_cLDR=True --loss_no_HDR=False --lr_G=0.00001
    %7 FAILED --model_name=FixUpUnet --mode_tonemap=LOG2  --batch_size=2 --loss_no_LDR=True --loss_no_cLDR=True --loss_no_HDR=False --lr_G=0.00001
    %8 --model_name=FixUpUnet --mode_tonemap=MIXED --batch_size=2 --loss_no_LDR=True --loss_no_cLDR=True --loss_no_HDR=False --lr_G=0.00001
    \bottomrule
    \end{tabular}

    \vspace{5pt}
    \caption{AllSky Ablation. Our HDR-losses improve retention of dynamic range.}
    \label{tab:AllSky_small}
    \begin{tabular}{@{}rcccccccccc@{}}
    \toprule
     Head & Losses & LDR $L_1$ $\downarrow$ & LDR $L_2$ $\downarrow$ & HDR $L_1$ $\downarrow$ & HDR $L_2$ $\downarrow$ & LPIPS $\downarrow$ & FID $\downarrow$ & EV ${I_f}/{I_r}$ & $\oiint_I$ ${I_f}/{I_r}$ \\
    \toprule
    RGB & LDR,cLDR     & 0.057 & 0.010 & 0.077 & 443.99 & 0.152 & 40.45 & 0.41 & 0.49 \\ %2 tanh
    RGB & LDR,cLDR,HDR & 0.057 & 0.009 & 0.077 & 443.77 & 0.146 & 41.27 & 0.71 & 0.51 \\ %1 tanh
    % RGB ReLU & LDR,cLDR     & 0.063 & 0.013 & 0.079 & 443.65 & 0.153 & 38.54 & 0.70 & \textbf{0.57} \\ %4 relu
    % RGB ReLU & LDR,cLDR,HDR & 0.070 & 0.015 & 0.081 & 442.65 & 0.156 & 37.83 & 0.82 & 0.55 \\ %3 relu
    
    \midrule
    %5,8,11 failed. --use_skip=False --use_priori=False
    ldr2EDR-9 & HDR & - & - & 0.09 & 533 & 0.159 & 57.3 & 0.7 & \textbf{0.53} \\ %9 --use_skip 9layers
    ldr2EDR-12 & HDR & - & - & 0.09 & 532 & 0.158 & 58.4 & 0.46 & 0.45 \\ %12 --use_skip 12layers
    ldr2EDR-9P & HDR & - & - & 0.09 & 532 & 0.158 & 46.6 & 0.85 & 0.48 &  \\ %10 --use_skip --use_priori 9layers
    ldr2EDR-12P & HDR & - & - & 0.09 & 533 & 0.181 & 47.8 & 0.52 & 0.45 &  \\ %13 --use_skip --use_priori 12layer
    \midrule
    latent2EDR-9 & HDR & - & - & 0.10 & 534 & 0.21 & 101 & - & 0.36 \\ %18 --use_skip 9
    latent2EDR-12 & HDR & - & - & 0.10 & 536 & 0.17 & 59 & \textbf{0.92} & \textbf{0.53} \\ %21 --use_skip   12
    latent2EDR-9P & HDR & - & - & 0.10 & 517 & 0.17 & 54 & 1.12 & 0.46 \\ %19 --use_skip --use_priori 9
    latent2EDR-12P & HDR & - & - & 0.10 & 525 & 0.17 & 55 & 0.89 & 0.46 \\ %22 --use_skip --use_priori 12
    % \midrule
    % latent2EDR-12PF & HDR & - & - &  &  &  &  &  &  \\
    % latent2EDR-12PF & HDR & - & - &  &  &  &  &  &  \\
    \bottomrule
    \end{tabular}
\end{table*}

\subsubsection{DeepClouds}

We summarize the results of our experimentation in \cref{tab:clearSky} and \cref{fig:X_DeepClouds_imagery}. 
We find little quantizeable difference between \textit{DeepClouds} and \textit{DeepClouds w/o Clear-Sky} for visual, illumination or DR metrics.
Though the Ho\v{s}ek-Wilkie provides intuitive control over clear-sky parameters, such as turbidity and ground albedo, DeepClouds keeps these parameters constant and after histogram equalization, retains no correlation to these parameters. 

% \begin{table*}[ht]
%   \caption{ClearSky Priori}
%   \label{tab:clearSky}
%   \begin{tabular}{@{}rcccccccccc@{}}
%   \toprule
%     & LDR $L_1$ $\downarrow$ & LDR $L_2$ $\downarrow$ & HDR $L_1$ $\downarrow$ & HDR $L_2$ $\downarrow$ & LPIPS $\downarrow$ & CLIP-IQA $\uparrow$ & DR ${I_f}/{I_r}$ & $\oiint_I$ ${I_f}/{I_r}$ \\
%     \toprule
%      DeepClouds               & 0.013 & 0.0008 & 0.07 & 412 & 0.21 & 0.54 & 0.36 & 0.59 \\
%      DeepClouds w/ Sun        & 0.019 & 0.006  & inf  & inf & 0.21 & 0.53 & 0.36  & $3e^{31}$ \\
%      DeepClouds w/o Clear-Sky & 0.016 & 0.0012 & 0.07 & 417 & 0.21 & 0.54 & 0.36 & 0.62 \\
%   \bottomrule
% \end{tabular}
% \end{table*}

The aggregation of the clear-sky sun in \textit{DeepClouds w/ Sun} significantly improves dynamic range, but visually in Fig.\cref{fig:X_DeepClouds_imagery} the sun appears extraterrestrial with a tiny solar disc. 
In terrestrial imagery, the sun is attenuated by the atmosphere and diffused through cloud formations and, though seamless in some, the aggregated sun is often visually unappealing and piercing cloud formations.
The increased $L_1$, $L_2$, and $\oiint_I$ is partially due to label misalignment, but primarily the sun-pass-through mechanism where, for a given sun ($S$), $I[|S|>10]' = I[|S|>10] + S[|S|>10]$. 
As a result, the model's generated $4.5EV$ sun is summed with a clear-sky sun to produce a cumulative $14EV$. 

%\begin{figure}[h]
%    \centering
%    \includegraphics[width=\linewidth]{figures/4_grid_DeepClouds}
%    \caption{HDRI Generated by DeepClouds Baselines}
%    \label{fig:3_DeepClouds_imagery}
%\end{figure}

\subsubsection{Text2Light}

\begin{figure}[ht]
    \centering
    \includegraphics[width=0.75\linewidth,keepaspectratio]{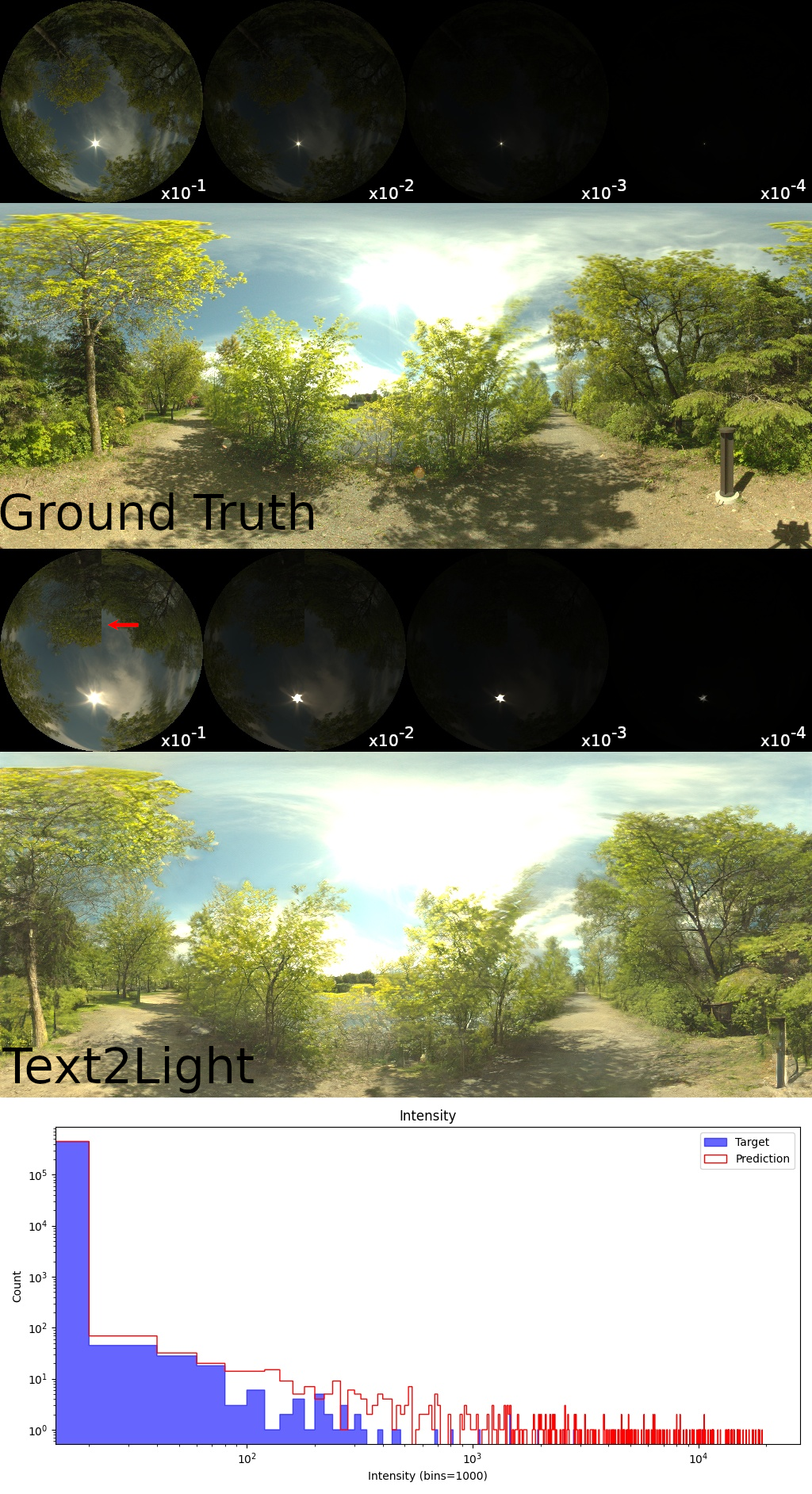}
    \caption{Text2Light: `purple petaled flower'. \textcolor{red}{Arrow} indicates latlong environment map seam.}
    \label{fig:3_Text2Light}
    \vspace{-0.5cm}
\end{figure}

As shown in \cref{fig:3_Text2Light} and \cref{tab:text2light}, Text2Light's iTMO and the subsequent exponential decompression function recreate EDR environment maps but over-expose the skydome and overshoot target illumination.
As a side effect of conditioning on latlong environment maps, the reconstructed skyangular environment maps exhibits a strong `seam' as shown in \cref{fig:3_Text2Light}. 
Of the 80 generated images with identified pairs in LOD it was observed that 17 images were unique, and the remaining 63 matched imagery within the generated subset between 2 and 12 times.
\Cref{fig:X_Text2Light_duplicates} illustrates 8 prompts which generated visual identical images.

% [brown tree trunks lot];9C4A7199
% [closeup photo of green leafed plants];9C4A7199
% [green and white leafed plants];9C4A7199
% [green leafed plants during daytime];9C4A7199
% [green leaves tree under blue sky during golden hour];9C4A7199
% [green-leafed trees near mountain at daytime];9C4A7199
% [orange petaled flowers near green trees at daytime];9C4A7199
% [purple petaled flower];9C4A7199

% \begin{figure}[h]
%     \centering
%     \includegraphics[width=\linewidth]{figures/9C4A7199_matches.png}
%     \caption{Text2Light: Visual matches for prompt `purple petaled flower'}
%     \label{fig:3_Text2Light_duplicates}
% \end{figure}

% \begin{table}
%   \caption{Text2Light. LOD ground truth CLIP-IQA is 0.38.}
%   \label{tab:text2light}
%   \begin{tabular}{@{}rcc@{}}
%     \toprule
%     & LDR & HDR \\
%     \midrule
%      $L_{1\downarrow}$  & 0.16 & 1.39 \\
%      DR ${I_f}/{I_r}$  & NA & 2.42 \\
%      II ${I_f}/{I_r}$  & 0.71 & 3.55 \\
%      LPIPS$_\downarrow$ & 0.44 & 0.42 \\
%      CLIP-IQA$_\uparrow$ & 0.39 & 0.52 \\
%      UIQI$_\uparrow$ & 0.026 & 0.022 \\
%     % & & & \\
%   \bottomrule
% \end{tabular}
% \end{table}

\begin{table}
    \caption{Text2Light. LOD ground truth CLIP-IQA is 0.38.}
    \label{tab:text2light}
    \begin{tabular}{@{}rcccccc@{}}
    \toprule
    & $L_{1\downarrow}$  & EV ${I_f}/{I_r}$ & $\oiint_I$ ${I_f}/{I_r}$ & LPIPS$_\downarrow$ & CLIP-IQA$_\uparrow$ & UIQI$_\uparrow$ \\
    \midrule
    LDR & 0.16 & - & 0.71 & 0.44 & 0.39 & 0.026 \\
    HDR & 1.39 & 2.42 & 3.55 & 0.42 & 0.52 & 0.022 \\
    \bottomrule
    \end{tabular}
    \vspace{-0.3cm}
\end{table}

% Images HDR 
% L1: 1.3911572694778442
% L2: 81957.0078125
% DR: 2.419948101043701
% II: 3.548002243041992
% EMD: 3.9262123107910156
% CLIP_IQA real: 0.3784251809120178
% CLIP_IQA fake: 0.5161558389663696
% UQI: 0.022444427013397217
% LPIPS: 0.41522812843322754

% Images LDR 
% L1: 0.15929968655109406
% L2: 0.05179284140467644
% DR: nan
% II: 0.7055259943008423
% EMD: 0.0
% CLIP_IQA real: 0.3784251809120178
% CLIP_IQA fake: 0.3859398066997528
% UQI: 0.0258657094091177
% LPIPS: 0.43556585907936096

% Images HDR (Unique)
% L1: 1.211929440498352
% L2: 14834.482421875
% DR: 1.9688328504562378
% II: 3.1944096088409424
% EMD: 3.2205164432525635
% CLIP_IQA real: 0.39162954688072205
% CLIP_IQA fake: 0.5242627859115601
% UQI: 0.027330411598086357
% LPIPS: 0.41108813881874084

%% Images LDR (Unique)
% L1: 0.16058699786663055
% L2: 0.05031457543373108
% DR: nan
% II: 0.7342574596405029
% EMD: 0.0
% CLIP_IQA real: 0.39162954688072205
% CLIP_IQA fake: 0.41317060589790344
% UQI: 0.031257178634405136
% LPIPS: 0.4367830753326416
\subsubsection{Tonemappers}

We ablate to determine the impact of tonemapper selection on dynamic range by training our baseline with $\gamma$, $Log_2$ and $\mu\text{-lawLog}_2$ tone-mappers. 
Each model was trained for 112 epochs at $512^2$ resolution with $lr=e^{-4}$ and an accumulated batch size of 64 using LDR and cLDR losses.
\Cref{tab:fixUpUnet_tonemappers} shows that aggressive $\mu\text{-lawLog}_2$ tonemapping retains cloud and skydome textures while fostering higher dynamic range (EV ${I_f}/{I_r}$) and illumination ($\oiint_I$ ${I_f}/{I_r}$). 

To demonstrate the contribution of our proposed losses, we train two additional models with $\mu\text{-lawLog}_2$ tonemapping and substitute the LDR losses for a global $L_1$ loss in LDR- and HDR- space respectively. 
As shown in \cref{tab:fixUpUnet_tonemappers}, both models exhibit lower dynamic range (EV ${I_f}/{I_r}$) and illumination ($\oiint_I$ ${I_f}/{I_r}$).

\subsubsection{AllSky}

To evaluate AllSky, we trained small $64^2$ models with $lr=e^{-4}$ for 200 epochs on 5,000 images, with baseline (RGB), \textit{ldr2EDR} and \textit{latent2EDR} head configurations.
We train the baseline model (RGB) with LDR, cLDR and optionally HDR losses and, where applicable, use our $\mu\text{-lawLog}_2$ tone-mapping operator.
As demonstrated in \cref{tab:AllSky_small}, including our HDR-losses improves illumination and significantly improves dynamic range.

For \textit{ldr2EDR} and \textit{latent2EDR} configurations, we ablate the number of hidden layers (9 or 12), and the inclusion of the $X,Y,Z,\Omega,Label$ priori channels (P).
\Cref{tab:AllSky_small} illustrates both configurations offer a notable increase in dynamic range comparable or better than the baseline (RGB) with our HDR-losses. 
The results show that 9 hidden layers with priori included produce the best \textit{ldr2EDR} and \textit{latent2EDR} performance for dynamic range (EV ${I_f}/{I_r}$) with minimal impact to visual quality metrics. 

\section{Discussion}

The proposed method improves on the current state-of-the-art but is not without limitation. 
Through \cref{tab:clearSky,tab:fixUpUnet_tonemappers,tab:AllSky_small}, we demonstrate that conventional metrics (e.g.\ $L_1$ and/or $L_2$) in LDR and HDR-space are grossly inadequate to evaluate the environment maps generated by DNN sky-models. 
As demonstrated, our proposed metrics for dynamic range (EV) and illumination ($\oiint_I$) offer greater sensitivity and are more indicative of the overall quality of an environment map towards downstream applications such as IBL. 

In this work, we demonstrate that an ANN can replace conventional tonemapping operators and their hybrid composites.
Where many of these tonemapping operators are color agnostic, we make the assumption that latent color space is entangled and cannot be independently processed.
We demonstrate this approach can be used to extend the dynamic range of sky-models with negligible impact to visual quality. 

Visual losses implemented in this work are limited to facilitate focus on dynamic range and illumination. 
Though our quantitative evaluation demonstrates improved dynamic range, visual results in \cref{fig:X_AllSky,fig:X_DeepClouds_imagery,fig:X_Text2Light_duplicates} demonstrate that cloud textures are essentially lost for all DNN sky-models. 
Further work is required as the functionality of visual losses (e.g.\ LPIPS) in alternative LDR-color-spaces is unstudied.

Issues also remain with cloud segmentation, where thresholding by color ratio can be demonstrated to be inconsistent with changes in illumination and seasonality \cite{koehler1991status}. 
As sky-models achieve higher photorealism and resolution, the limitations of existing dataset will become apparent. 
Few EDR sky imagery datasets exist and HDRDB at higher resolutions is plagued by HDR artifacts including significant ghosting. 

\section{Conclusions}

In this work, we propose a Deep Neural Network (DNN) leveraging physically captured HDRI with the objective of learning physically accurate skies with diverse and photorealistic atmospheric formations.
We demonstrate that our model (AllSky) generates global representations of physically captured environment maps per user-controlled label maps, with improved retention of the sky's Extended Dynamic Range (EDR).
We further showcase the current shortfalls of DNN sky-models (ours-included) in their capacities to produce photorealistic weathered skies with faithful and controllable HDR illumination.
We demonstrate the current evaluation of sky-models is inadequate for demonstrating the quality of generated environment maps towards downstream applications such as Image Based Lighting (IBL).
Significant work remains for DNN sky-models to become interchangeable with parametric sky-models and comparable in quality to physically captured HDRI.

%%%%%%%%% Bibliography %%%%%%%%%
% \cite{*}
{
    \small
    \bibliographystyle{ACM-Reference-Format}
    \bibliography{main}
}

%%%%%%%%% Appendix %%%%%%%%%
\clearpage
\appendix
\onecolumn
\section{Figures}
\begin{figure*}[hbt!]
    \includegraphics[width=0.9\textwidth,keepaspectratio]{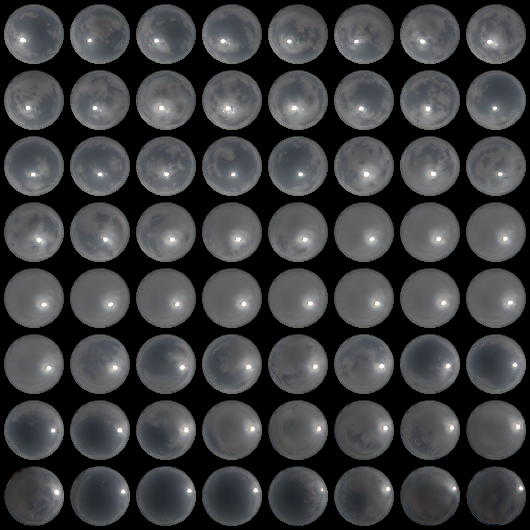}
    \caption{Images from modular training of \textit{latent2EDR}}
    \label{fig:X_AllSky}
\end{figure*}

\begin{figure*}[hbt!]
\includegraphics[width=.9\textwidth,keepaspectratio]{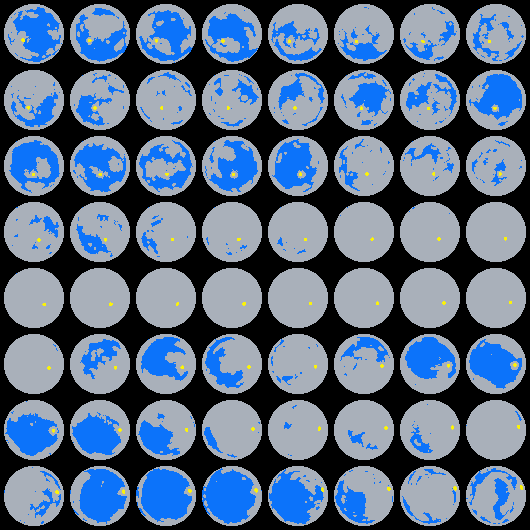}
    \caption{HDRDB labels paired to \cref{fig:X_AllSky}}
    \label{fig:3_AllSky_Imagery}
\end{figure*}
\clearpage
\section{Text2Light}

Quantitative evaluation of Text2Light was completed via unaltered implementation the author's published model and associated checkpoints \cite{text2light}.
For this work, Text2Light's \textit{outdoor} functionality was evaluated using the author's published list of textual CLIP prompts, matching generated images to ground truth images from the Laval Outdoor dataset.
Image pairs were identified via FLANN matched ORB key-points. Given the visual similarity and negligible misalignment, further pairs were identified via the author's scale invariant loss.
\centerline{
\rotatebox{90}{
    \begin{minipage}{0.87\textheight}
        \centering
        \includegraphics[width=\textwidth,keepaspectratio]{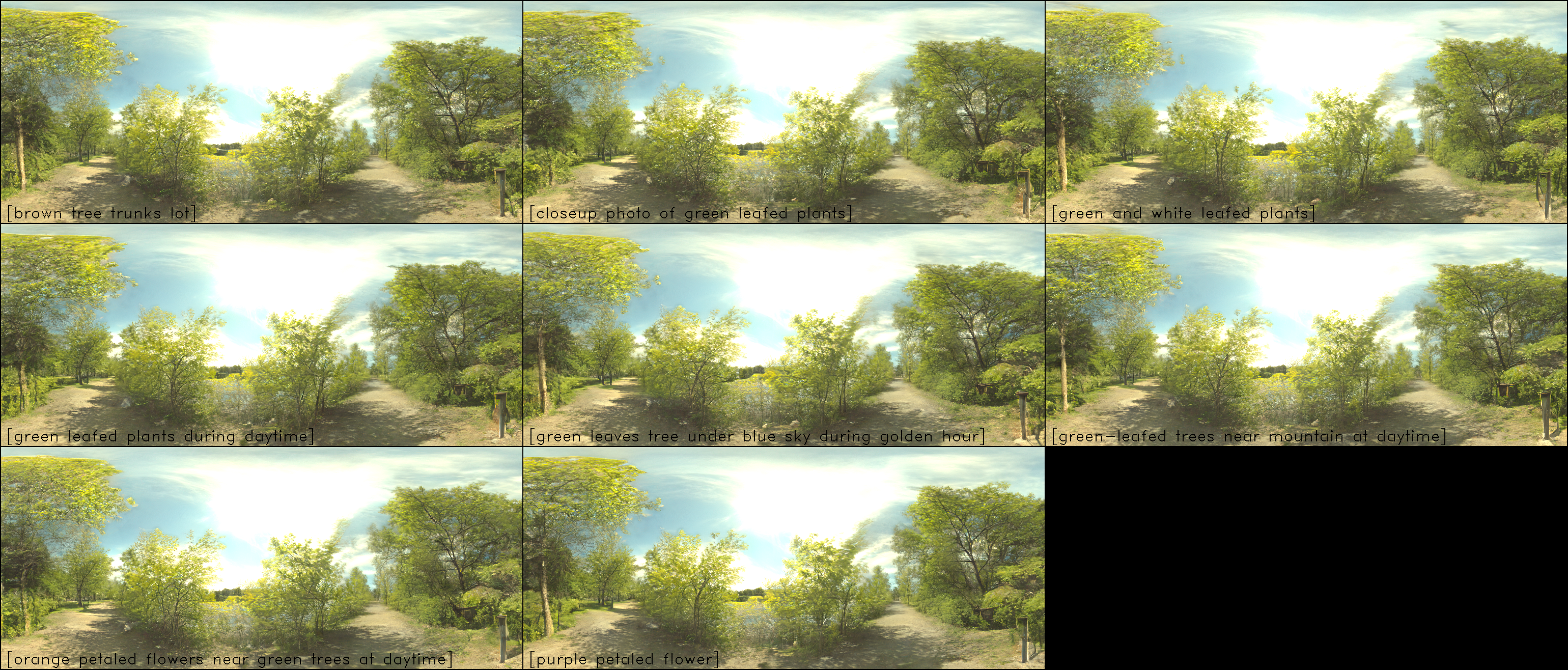}
        \captionof{figure}{Text2Light: Visual matches for prompt `purple petaled flower'}
        \label{fig:X_Text2Light_duplicates}
    \end{minipage}
}
}
\clearpage
\section{DeepClouds}

For evaluation purposes, DeepClouds was reproduced per source-code provided by the authors with model and parameters unchanged \cite{DEEPCLOUDS_22}.
A pytorch-lightning framework to achieve the required prolonged training on our larger HDRDB dataset.

\centerline{
\rotatebox{90}{
    \begin{minipage}{0.92\textheight}
        \centering
        \includegraphics[width=\textwidth,keepaspectratio]{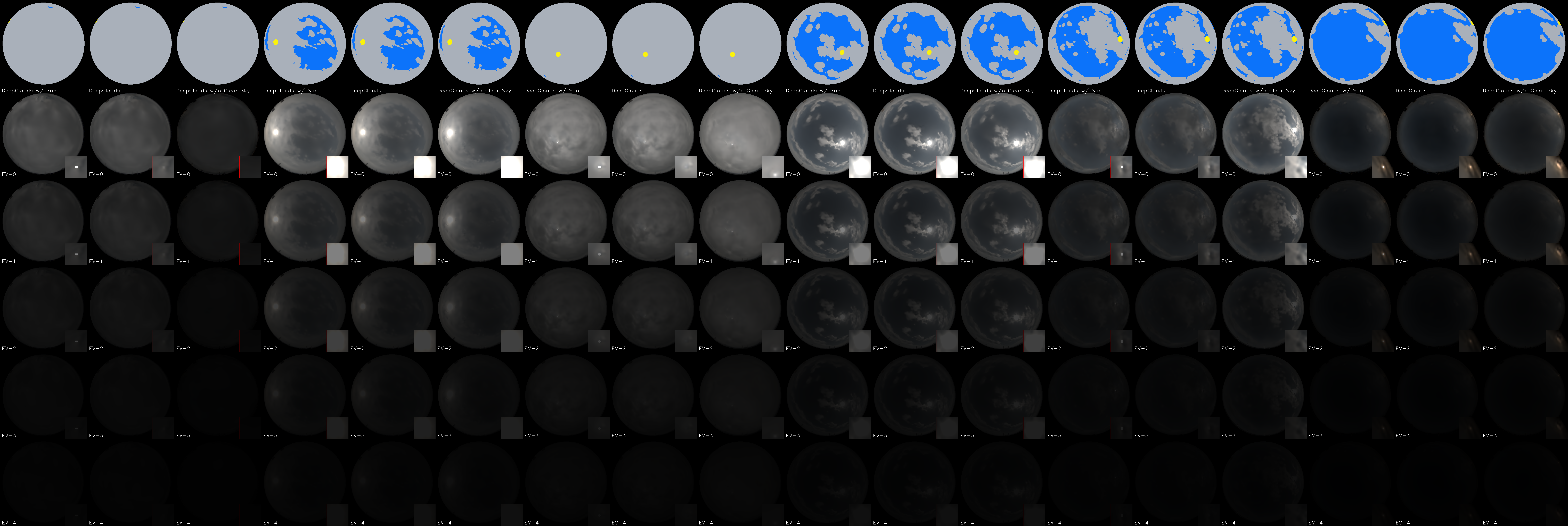}
        \captionof{figure}{HDRI Generated by DeepClouds Baselines}
        \label{fig:X_DeepClouds_imagery}
    \end{minipage}
}
}

%%%%%%%%% END DOCUMENT %%%%%%%%%
\end{document}